\documentclass[11pt]{article}
\usepackage{histocryptstyle}
\usepackage{mathptmx}
\usepackage{url}
\usepackage{graphicx}
\usepackage{multirow}
\usepackage{latexsym}
\usepackage{amsmath}
\usepackage{blindtext}
\usepackage{xcolor}

\title{Joint Transcription and Decryption of Images of Encrypted Handwritten Documents: A Comparison with the Traditional Pipeline}

\author{
\textbf{Marino Oliveros-Blanco} \\
\textbf{Lei Kang} \\
\textbf{Alicia Forn\'{e}s} \\
Computer Vision Center \\
Department of Computer Science \\
Universitat Aut\`{o}noma de Barcelona, Spain \\
{\ttfamily marino.oliverosblanco@gmail.com} \\
{\ttfamily \{lkang,afornes\}@cvc.uab.es}
\And
\textbf{Be\'{a}ta Megyesi} \\
Department of Linguistics \\
Stockholm University, Sweden \\
{\ttfamily beata.megyesi@ling.su.se}
}
\date{}
\begin{document}

\maketitle

\begin{abstract}
  Historical encrypted manuscripts present a  challenging problem at the intersection of cryptology, linguistics, paleography, and computer vision. Current automatic decipherment approaches usually rely on a two-stage pipeline: transcription of cipher symbols from manuscript images, followed by decryption into plaintext. However, this design is sensitive to transcription errors, which propagate to the final output.  We present Direct Image Decryption, an end-to-end approach that directly maps encrypted manuscript images to plaintext, bypassing the intermediate transcription stage. Using the Copiale cipher as a case study, we build a synthetic data generation pipeline to create large-scale cipher-like training data and compare the traditional pipeline with the proposed joint architecture. Results show that joint image-to-plaintext modeling is a promising alternative to traditional transcription-based pipelines.
\end{abstract}

\section{Introduction}

Historical ciphers constitute an important part of the written record of early modern and modern Europe, appearing in diplomatic correspondence, private letters, intelligence reports, and the papers of learned and secret societies. Many such documents remain only partially studied, not because they have been lost, but because their encrypted content is still difficult to access. Their analysis requires expertise from several fields, including cryptology, linguistics, paleography, history, and, increasingly, computer vision. 

In practice, the decipherment of encrypted manuscripts usually follows a two-stage process. First, the cipher symbols are transcribed from the image into a textual representation. Second, the resulting sequence is subjected to cryptanalytic and linguistic analysis to recover the plaintext and reconstruct the underlying encryption system. However, this sequential pipeline has important limitations: Errors introduced during transcription propagate directly to the decryption stage, compromising the final output. In addition, transcription often requires considerable manual effort and specialist knowledge, which restricts the scalability of existing approaches.

Motivated by these limitations, this work investigates an alternative approach, which we call \textit{Direct Image Decryption}. Rather than treating transcription and decryption as separate tasks, this approach learns a direct mapping from images of encrypted handwritten text to decrypted plaintext within a single end-to-end model. The goal is to reduce the impact of transcription errors, alleviate the manual bottleneck associated with symbol transcription, and exploit visual information that may be lost when manuscript images are first converted into discrete symbol sequences. To examine these possibilities, we implement and compare two deep learning architectures: a traditional two-stage transcription--decryption pipeline and a joint image-to-plaintext model.

We evaluate both approaches on real and synthetically generated data. As a case study, we focus on the well-known Copiale cipher. This 18th-century manuscript, discovered in Germany, consists of 105 pages written in a large set of symbols and abstract glyphs. The system is a homophonic substitution cipher in which individual plaintext letters are represented by multiple cipher symbols drawn from an alphabet of approximately 100 distinct glyphs. The manuscript was deciphered in 2011 \cite{Knight:11}, showing that the text encoded German and described rituals associated with Freemasonry and a secret society known as the \emph{Oculists}. 

It must be noted that our model is trained to decrypt this specific substitution system rather than to perform fully cipher-agnostic decryption. However, our framework could be applied to other encrypted handwritten sources.

\section{Related Work}
Work on historical ciphers has increasingly combined traditional cryptanalytic scholarship with computational methods. In case of historical encrypted manuscripts, this often implies the transcription of cipher symbols from document images, and the subsequent decipherment of the resulting symbol sequences.

An example in computational historical cryptology is the decipherment of the Copiale manuscript by \newcite{Knight:11}, which showed how large historical ciphers can be approached through a combination of transcription, statistical language modeling, clustering, and cryptanalytic analysis. Similar workflows have also been applied to other historical ciphers. At the same time, these studies make clear that transcription remains a substantial practical challenge. For Copiale, manual transcription reportedly required around 35 minutes per page once the symbol inventory had been identified, and considerably longer otherwise \cite{Knight:11}. \newcite{Dinnissen:21} likewise describe the significant effort required to transcribe the Ramanacoil manuscript before any decipherment could be attempted. This dependence on high-quality transcription continues to shape the feasibility of computational work on historical ciphers.

On the decipherment side, there was a growing interest in neural approaches. \newcite{Kambhatla:18} introduced neural language-model-based methods with beam search for cipher decipherment. \newcite{Aldarrab:21} proposed transformer-based multilingual models that achieved strong results on both synthetic and handwritten ciphers. \newcite{Aldarrab:22} further explored neural end-to-end settings and highlighted the continuing difficulty of transferring models trained on synthetic data to real manuscripts.

In parallel, advances in handwritten text recognition have influenced the treatment of encrypted manuscripts. Convolutional Recurrent Neural Networks (CRNNs) trained with Connectionist Temporal Classification (CTC) loss \cite{Graves:06} have become a standard architecture for sequence recognition from images \cite{Shi:17}. They eliminate the need for explicit character segmentation, useful when the segmentation of symbols is difficult. \newcite{Yin:19} addressed automatic segmentation, glyph recognition and transcription for historical ciphers, emphasizing the challenges posed by non-standard alphabets, degraded images, and the absence of lexical constraints such as dictionary-based error correction. \newcite{Bluche:17} introduced an end-to-end model combining LSTMs with attention mechanisms for paragraph-level recognition.
More recently, the ICDAR Competition on Handwriting Recognition of Historical Ciphers \cite{Fornes:24} has provided standardized shared datasets and evaluation protocols, enabling more systematic comparison of methods.

Despite the progress, current neural decryption methods still depend fundamentally on accurate transcription, which creates error propagation that directly affects decryption performance. More broadly, most approaches continue to treat transcription and decipherment as separate tasks. As a result, overall system performance remains closely tied to transcription quality, while the scarcity of annotated historical material continues to encourage heavy reliance on synthetic data.

We believe three fundamental limitations of current approaches persist:

\textbf{1. Error propagation:} Errors from transcription to decryption compound inaccuracies and limit overall system performance. A single transcription error can cascade through the decryption process and corrupt surrounding text interpretation.

\textbf{2. Transcription bottleneck:} This process requires substantial manual effort or large quantities of training data. For many historical ciphers, neither sufficient manpower nor annotated data is readily available. The scarcity of real manuscript data, rarely exceeding a few thousand lines, necessitates heavy reliance on synthetic data generation, which often fails to capture authentic manuscript complexity. This limitation makes decrypting documents such as the Voynich Manuscript \cite{Clemens:16}, part of the Zodiac Killer Ciphers (particularly the brief Z13 and Z32), or the Beale Ciphers \cite{Gillogly:80} particularly challenging.

\textbf{3. Synthetic-to-real gap:} While synthetic data enables training of large-scale models, it typically fails to replicate exact characteristics of historical manuscripts, including aging effects, ink degradation, writing style variations, and document damage. As a result, models trained primarily on synthetic data often exhibit degraded performance on original historical ciphers.

These limitations motivate our Direct Image Decryption approach, which bypasses transcription entirely and directly maps visual features to decrypted plaintext, representing a fundamental departure from established methodologies.

\section{Synthetic Data Generation}
The scarcity of annotated historical encrypted manuscripts presents the most fundamental challenge for training deep learning models. With only approximately 2,000 segmented line images available from the original Copiale manuscript \cite{Fornes:24}, we developed a comprehensive synthetic data generation pipeline to produce training samples of sufficient quantity and quality for robust model development of both the transcription-decryption pipeline and the Direct Image Decryption model. This pipeline takes lines of text as input and generates augmented images of the original text encoded into Copiale, along with the transcriptions and original decrypted plaintext. 

The present work builds on these developments by exploring a joint formulation in which encrypted manuscript images are mapped directly to plaintext, allowing transcription and decryption to be learned within a single model.

Our synthetic data must satisfy three requirements: visual similarity to the Copiale manuscript, including appropriate symbol shapes, spacing, and overall appearance; linguistic patterns reflecting 18th-century German, as this was the language encoded in the original cipher; and realistic degradation effects—including noise, ink variations, and aging marks—to reduce the gap between synthetic samples and authentic historical documents.

\subsection{Text Source Selection}

We selected historical German texts chronologically and stylistically aligned with Copiale. Our primary corpus contains four major works: Goethe's \textit{Faust} (1808-1832), Kant's \textit{Critique of Pure Reason} (1781), \textit{the Lutheran Bible} (1760 revision), and Adalbert Stifter's \textit{Nachsommer} (1857), providing over 115,000 lines of period-appropriate German text. The dataset generated from this corpus is called the ``Faust" dataset. 

Text preprocessing filtered the corpus to retain only the 106 characters present in the original cipher, removing modern punctuation and uncommon characters. Line segmentation ensured generated images contained 12-40 characters, matching the length distribution observed in the original manuscript.
For additional evaluation, we generated datasets using English texts (\textit{American Psycho} by Bret Easton Ellis, \textit{East of Eden} by John Steinbeck) and Old German poetry (\textit{Hymns of the Night} by Novalis, 1,300 images).

\subsection{Visual Representation and Augmentation}

The encoding employs the \textit{``Copiale.ttf"} font file, which maps standard Unicode characters to cipher glyphs, enabling automatic generation of cipher-like text from plaintext input. This encoding follows the fixed substitution key of the Copiale cipher specifically; all models trained on this synthetic data learn to decrypt this particular key rather than performing general cipher-agnostic decryption. The mapping process utilizes the vocabulary file from the DECRYPT project \cite{Megyesi:20}, ensuring our synthetic data maintains the same symbol-to-meaning relationships as the authentic manuscript data.

\begin{figure}[ht]
\centering
\includegraphics[width=0.8\columnwidth]{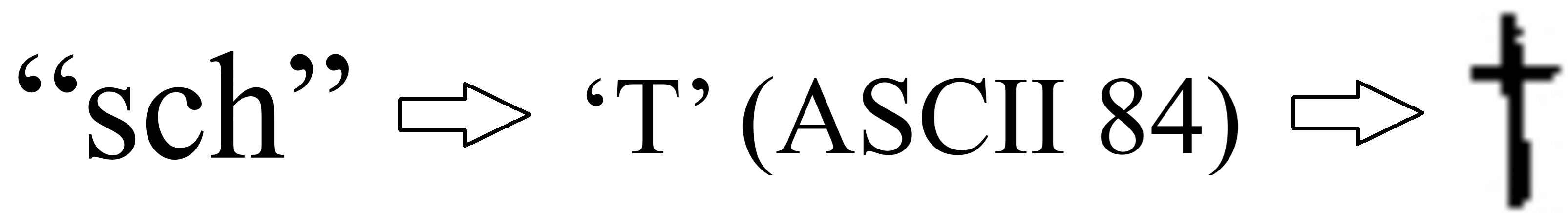}
\caption{Input to Copiale encoding}
\label{fig:encoding-example}
\end{figure}

Initial image generation produces clean renderings, just encoded text with an applied font. To obtain visual similarity to the aged manuscripts, we apply comprehensive augmentation: \textbf{Degradation effects} include Gaussian noise (paper texture, scanning artifacts), random erosion and dilation (ink spread, fading), gamma correction (brightness variations), and Kanungo noise patterns (spots, stains, fiber patterns). \textbf{Geometric transformations} apply random rotation ($\pm$3 degrees), shearing (perspective distortions), random scaling, and random cropping.

\begin{figure}[ht]
\centering
\includegraphics[width=1.0\columnwidth]{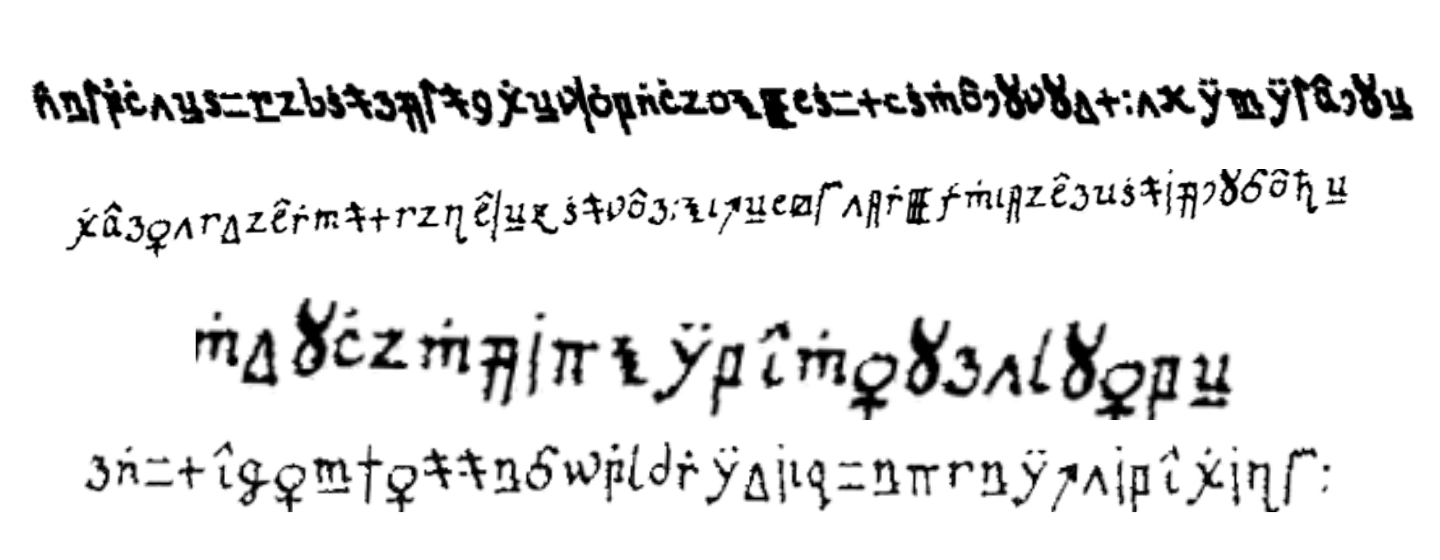}
\caption{Different augmentation effects}
\label{fig:augmentation-example}
\end{figure}

The augmentation parameters were carefully tuned to produce realistic variations as similar as possible to the original manuscript, balancing degradation intensity to avoid over-distortion while capturing authentic aging characteristics. Figure~\ref{fig:augmentation-example} shows the range of augmentation effects obtainable through our pipeline.

\begin{figure}[ht]
\centering
\includegraphics[width=1\columnwidth]{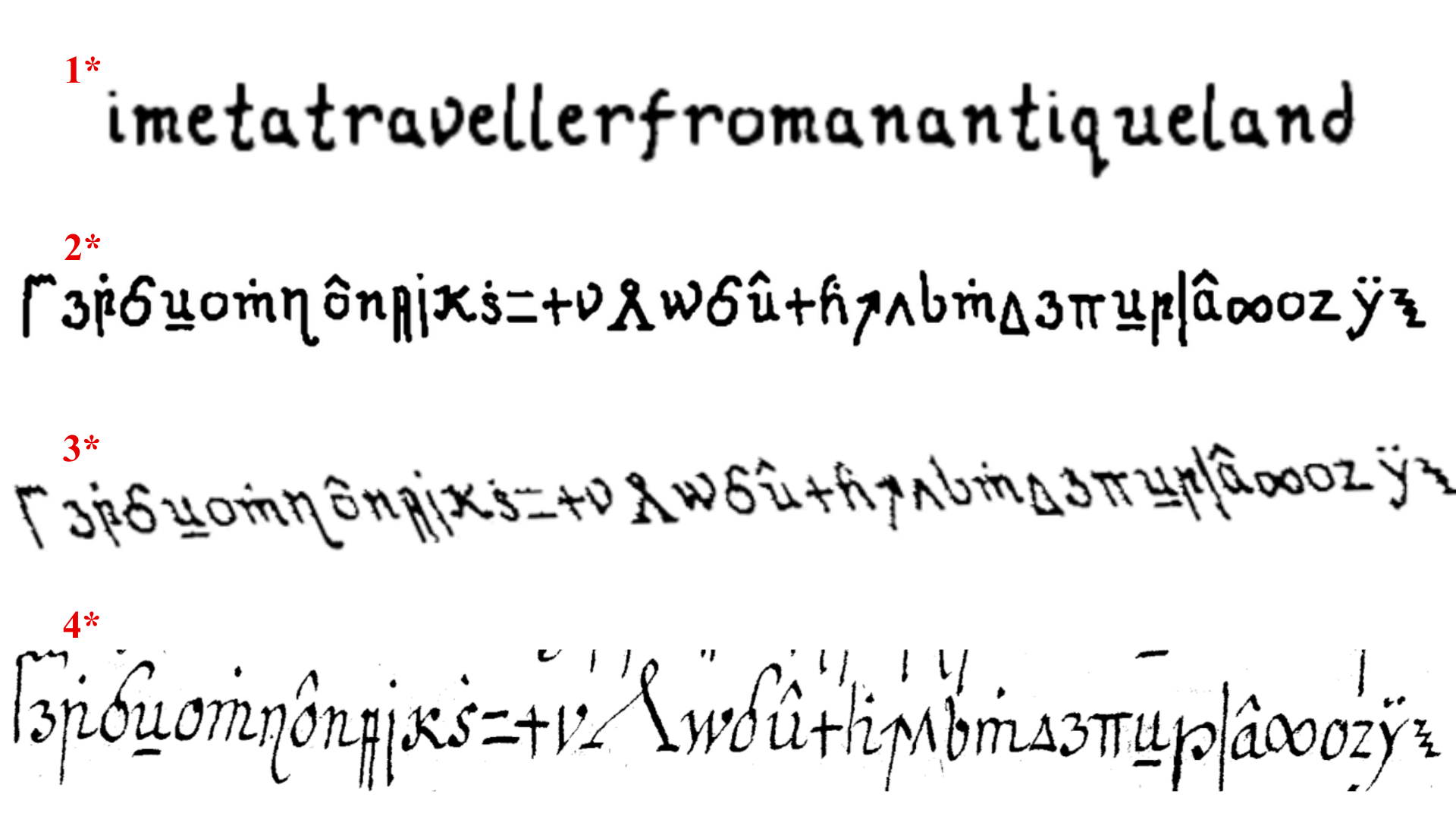}
\caption{Comparison of 1* Plaintext in Copiale font, 2* Encoded text (non-augmented), 3* Encoded text (augmented), and 4* Original manuscript image.}
\label{fig:text-stages-comparisons}
\end{figure}

\subsection{Dataset Statistics}

\begin{figure}[ht]
\centering
\includegraphics[width=1\columnwidth]{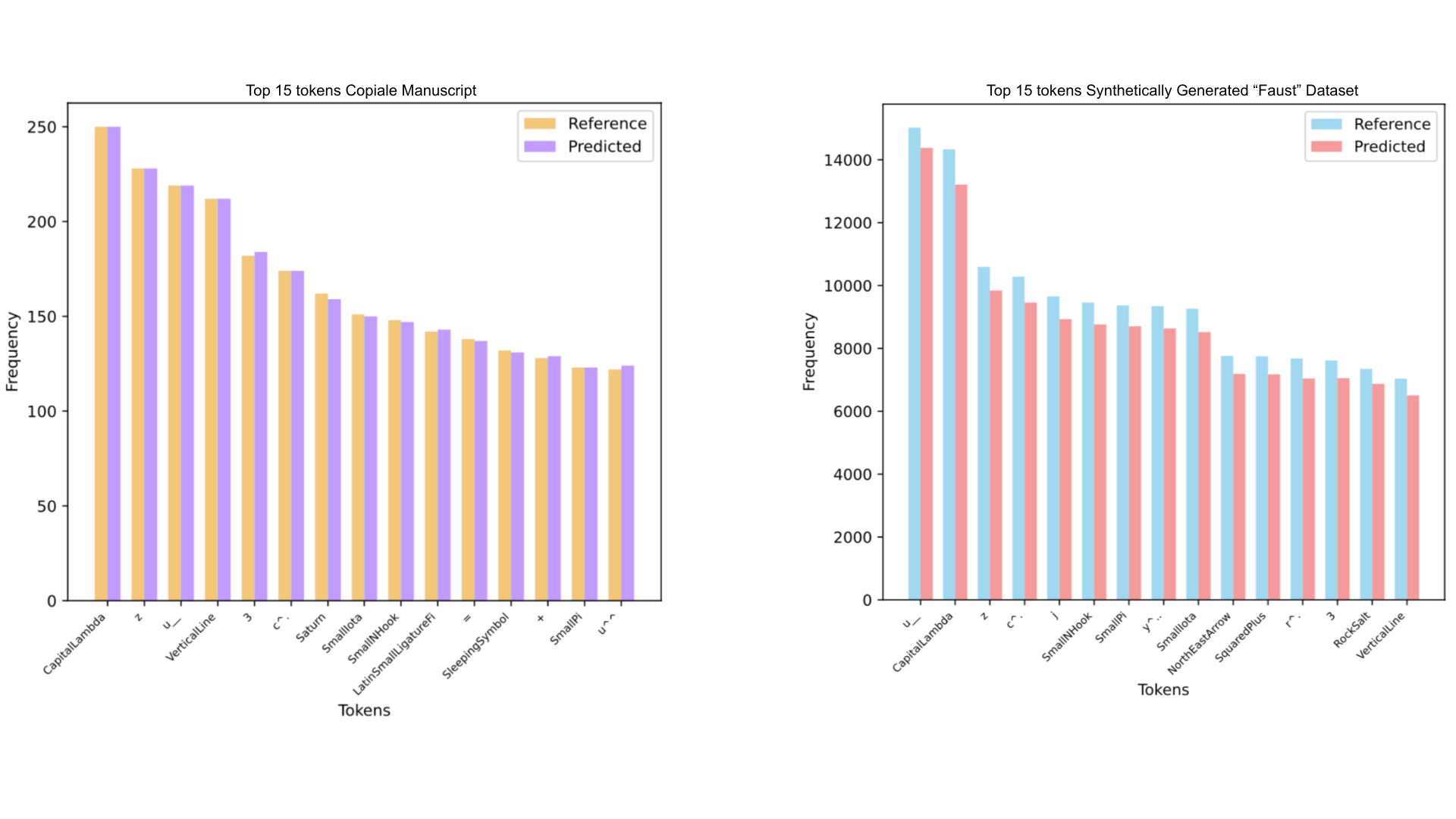}
\caption{Token frequency: Copiale dataset vs. Synthetically generated ``Faust'' dataset.}
\label{fig:frequency-token}
\end{figure}

Our primary synthetic dataset, ``Faust", comprises 115,000 line images with 80/10/10 split for training, validation, and testing. The Copiale dataset consists of 2,000 grayscale images with transcription and decrypted plaintext. Figure~\ref{fig:text-stages-comparisons} shows that augmentation effectively achieves correct symbol morphology, spacing patterns, and degradation effects; although real manuscripts exhibit more pronounced historical wear, and paper-like effects. Vocabulary distribution in our synthetic data closely resembles the original manuscript (Figure~\ref{fig:frequency-token}), ensuring models encounter symbol patterns consistent with real manuscript images.

\section{Transcription \& Decryption Pipeline}

\begin{figure}[ht]
\centering
\includegraphics[width=1\columnwidth]{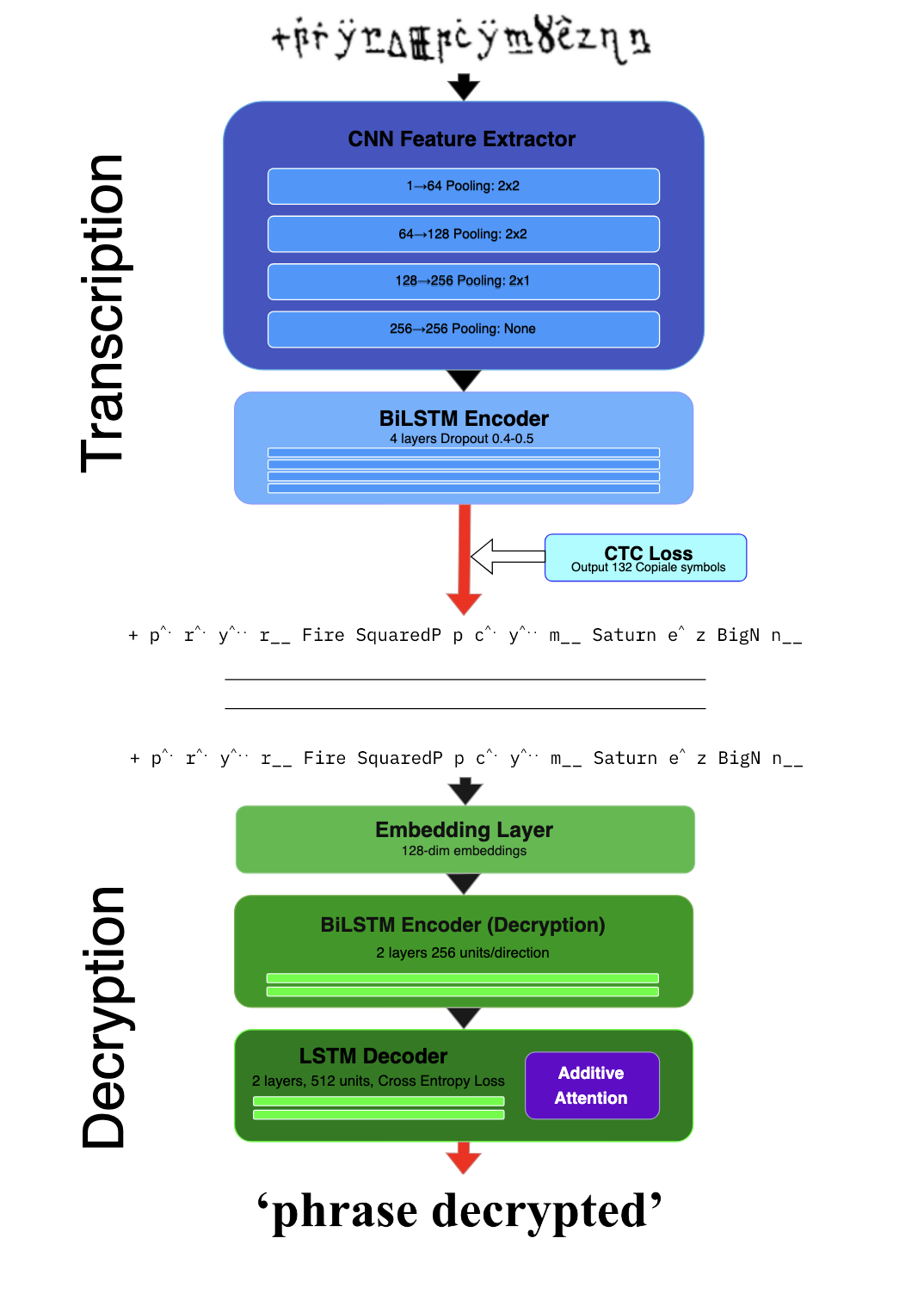}
\caption{Two-stage pipeline: image $\rightarrow$ transcription + transcription $\rightarrow$ decryption $\rightarrow$ plaintext.}
\label{fig:transcription-decryption}
\end{figure}

The two-stage pipeline represents the traditional approach to the decipherment of encrypted manuscripts, consisting of first transcription followed by decryption, as shown in Figure \ref{fig:transcription-decryption}.

\subsection{Stage 1: Transcription}

The transcription stage converts images of cipher manuscripts into sequences of symbol tokens. Obviously, achieving high-quality transcription is critical for the posterior decryption.

\begin{figure}[ht]
\centering
\includegraphics[width=1\columnwidth]{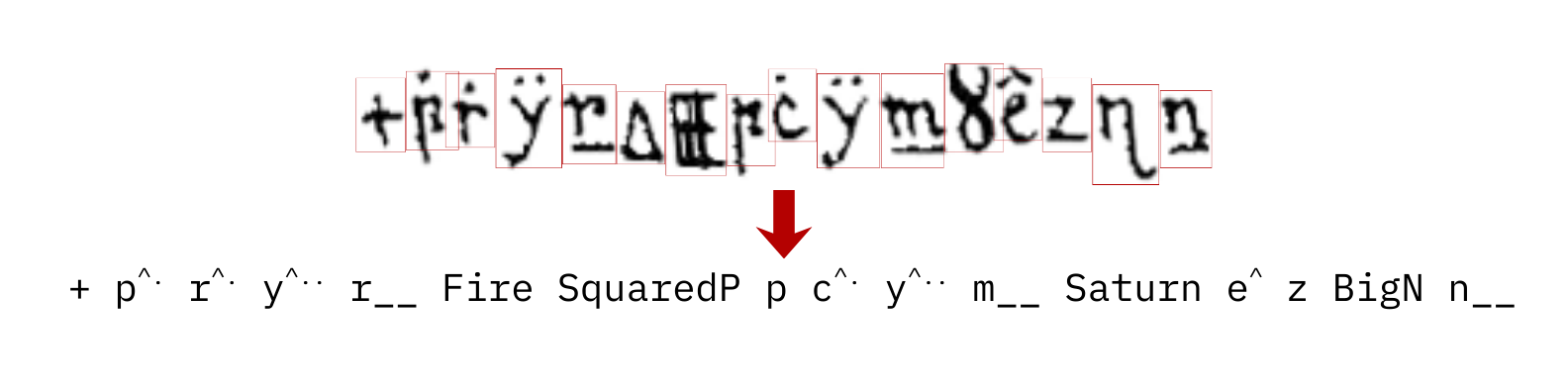}
\caption{Example of successful transcription, in which the  input image is converted into a cipher symbol token sequence.}
\label{fig:transcription-example}
\end{figure}

\subsubsection{Model Architecture}

The transcription model employs a Convolutional Recurrent Neural Network (CRNN) architecture with Connectionist Temporal Classification (CTC) loss \cite{Shi:17}. Input images are resized to 64 pixels height, padded/cropped to 800 pixels width, and normalized to [0,1].

\textbf{CNN Feature Extractor:} Four convolutional blocks (1$\rightarrow$64$\rightarrow$128$\rightarrow$256$\rightarrow$256 channels) progressively extract hierarchical features with 3$\times$3 convolutions, batch normalization, ReLU activation, and pooling (2$\times$2, 2$\times$2, 2$\times$1, none). Output feature maps are reshaped to (batch\_size, width/4, 2048).

\textbf{Bidirectional LSTM:} 4 layers with 256 hidden units per direction process CNN features bidirectionally to disambiguate visually similar symbols. Dropout (0.5) provides regularization, producing 512-dimensional vectors per timestep.

\textbf{CTC Classification:} The vocabulary comprises 132 Copiale cipher tokens plus special tokens (blank, padding, UNK). Greedy CTC decoding selects the most probable token at each timestep, then collapses repetitions and removes blanks.
The CTC loss \cite{Graves:06} enables alignment-free training by marginalizing over all possible alignments:
\begin{equation}
\mathcal{L}_{\text{CTC}} = -\log P(y|x) = -\log \sum_{\pi \in \mathcal{B}^{-1}(y)} \prod_{t=1}^{T} p_t(\pi_t|x)
\end{equation}
where $y$ is the target symbol sequence, $x$ is the input image, $T$ is the sequence length, $\pi$ is an alignment path, $\mathcal{B}^{-1}(y)$ is the set of valid alignments, and $p_t(\pi_t|x)$ is the probability of symbol $\pi_t$ at timestep $t$.

\subsubsection{Training Configuration}

AdamW optimizer \cite{Loshchilov:19} with learning rate 3$\times$10\textsuperscript{-4}, weight decay 1$\times$10\textsuperscript{-5}, batch size 8, gradient clipping (max norm 1.0), and ReduceLROnPlateau scheduler (factor 0.1, patience 5 epochs). Training runs 100 epochs with early stopping.

\subsection{Stage 2: Decryption}

The decryption stage consists of taking transcribed symbol sequences and generating decrypted German plaintext using a sequence-to-sequence architecture with attention mechanism.

\subsubsection{Model Architecture}
It consists of an Encoder-decoder architecture with additive attention. The encoder uses bidirectional LSTM (2 layers, 256 units per direction) to create contextual representations. The decoder generates plaintext character-by-character through matching LSTM architecture, with attention focusing on relevant encoded cipher portions. Character-level embeddings of dimension 128 for both input and output. Output projection maps decoder states to German alphabet vocabulary (uppercase, lowercase, special characters, control tokens: SOS, EOS, PAD, UNK). The decryption model uses the same vocabulary structure as transcription for its input, ensuring seamless integration with approximately 100-200 cipher symbol tokens depending on the dataset.

\subsubsection{Training Strategy}

AdamW optimizer (learning rate 1$\times$10\textsuperscript{-3}, weight decay 1$\times$10\textsuperscript{-4}, dropout 0.4), batch size 16, 15-35 epochs with early stopping based on validation edit distance. The decryption model is trained from scratch on synthetic data  with no external pretraining or pretrained language model weights. ReduceLROnPlateau scheduler reduces learning rate when validation plateaus. Teacher forcing is annealed linearly from 1.0 to 0.0 across epochs, and gradient clipping is applied at max norm 1.0.

\section{Direct Image Decryption}

Our joint architecture, namely, Direct Image Decryption, learns direct image-to-plaintext mapping in a single end-to-end model. By jointly optimizing visual feature extraction and decryption without intermediate discrete decisions, the model discovers which visual features are most relevant for decipherment, addressing the limitations outlined in Section 2: error propagation, information loss during symbolic conversion, and architectural complexity of maintaining separate models.

\subsection{Architecture Overview}

\begin{figure}[ht]
\centering
\includegraphics[width=1\columnwidth]{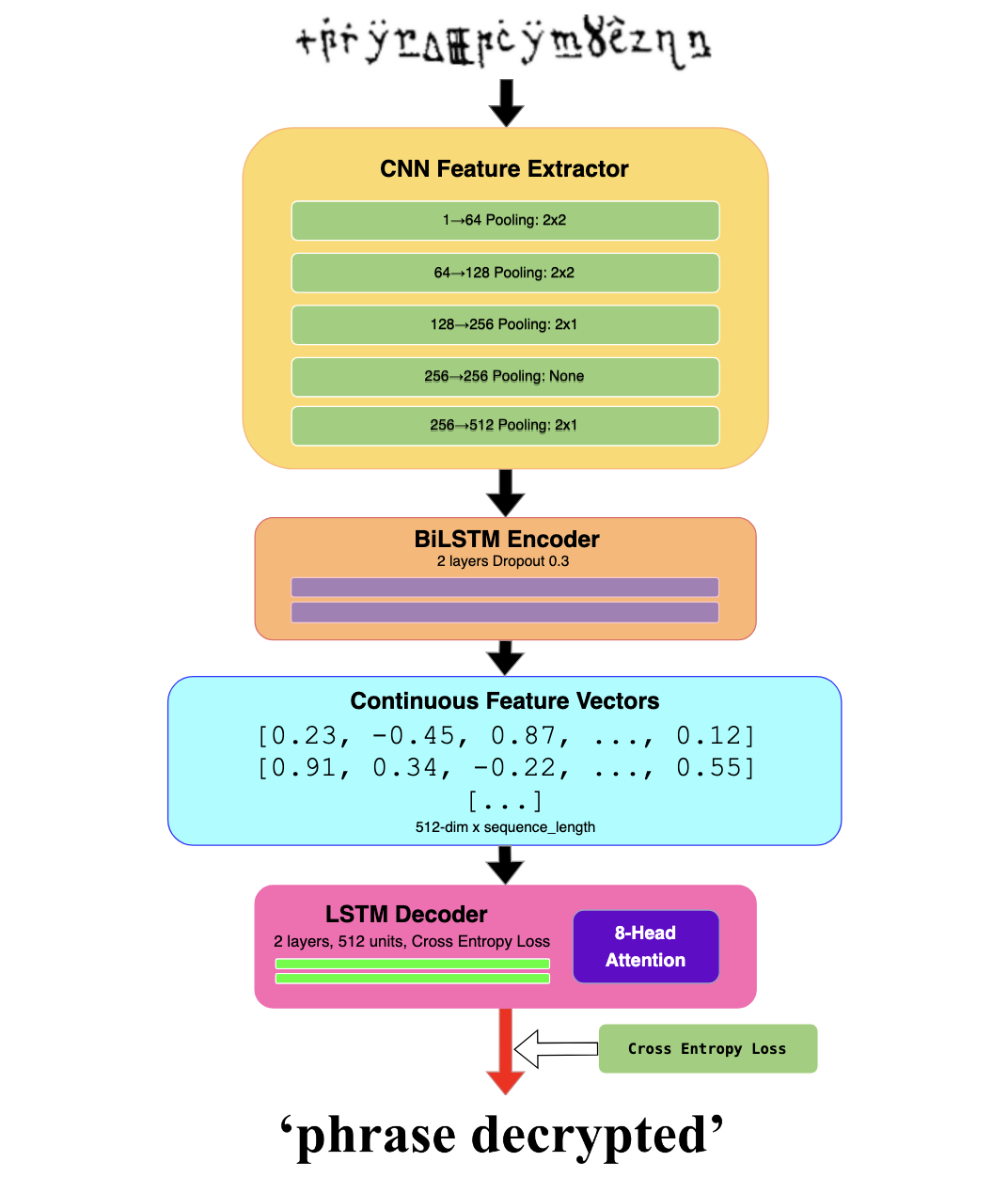}
\caption{Direct Image Decryption architecture: images processed through CRNN feature extraction, decoded directly to plaintext through attention-based LSTM decoder.}
\label{fig:pixel-hacking}
\end{figure}

The architecture comprises a CRNN feature extractor producing sequential visual representations, and an attention-based decoder generating plaintext characters autoregressively. The critical difference from the two-stage pipeline: intermediate representations remain continuous—the model never commits to discrete cipher symbol decisions, allowing end-to-end gradient flow and joint optimization.

\subsection{CRNN Feature Extractor}

The feature extractor extends the CRNN architecture from Section 4 with modifications to support end-to-end training. It processes grayscale images through a deeper five-block CNN structure (adding a fifth block: 256$\rightarrow$512 channels with 2$\times$1 pooling), expanding upon the four-block design used in transcription.

Following CNN blocks, feature maps of dimension (batch\_size, 512, height/16, width/4) are reshaped to dimension (batch\_size, width/4, 512 $\times$ height/16). A 2-layer bidirectional LSTM (256 units per direction) produces contextualized visual representations of dimension (batch\_size, sequence\_length, 512).
The CRNN can initialize with pretrained transcription model weights (Section 4.1.1), providing a strong starting point. However, unlike the transcription pipeline where CRNN weights remain fixed during decryption training, Direct Image Decryption allows fine-tuning during the end-to-end decryption, permitting optimization. The feature extractor output—sequences of 512-dimensional vectors—captures cipher symbols, spatial relationships, and visual characteristics without committing to discrete symbol decisions.

\subsection{Attention-Based Decoder}

The decoder generates German plaintext characters autoregressively, conditioning on encoded image features and previously generated characters. It employs 2-layer LSTM (512 hidden units) with 8-head multi-head attention mechanism.

At each timestep, the decoder receives embedding of the previous character (dimension 128) or SOS token. The LSTM produces a query vector, and multi-head attention (8 heads, per-head dimensionality 64) computes scores between query and encoded image features, determining which manuscript regions are most relevant for predicting the current character.

Attended features are concatenated with LSTM output, passed through linear projection (dimension 512), then final output projection to German alphabet vocabulary. Cross-entropy loss trains the model with gradients flowing backward through attention, decoder, and feature extractor. Teacher forcing during training stabilizes learning; inference uses autoregressive generation until an EOS token is found.

\subsection{Training Configuration}

End-to-end training uses AdamW~\cite{Loshchilov:19} (learning rate 1$\times$10\textsuperscript{-3}, weight decay 1$\times$10\textsuperscript{-4}, dropout 0.3), batch size 16, 35-50 epochs with early stopping based on validation edit distance. The CRNN encoder is initialized with weights from the pretrained transcription model (Section 4.1), then the full architecture is fine-tuned end-to-end; the attention decoder is always trained from scratch. Gradient clipping (max norm 1.0) prevents instability. ReduceLROnPlateau scheduler (factor 0.1, patience 5 epochs) enables fine-grained optimization.

The model is trained end-to-end to directly generate plaintext from images:

\begin{equation}
\mathcal{L}_{\text{DID}} = -\sum_{t=1}^{T} \log P(w_t | w_{<t}, I; \theta)
\end{equation}

where $I$ is the input manuscript image, $w_t$ is the plaintext character at position $t$, $w_{<t}$ represents all previous characters, and $\theta$ represents all model parameters. This objective directly optimizes for accurate plaintext generation without intermediate transcription objectives, allowing gradients to flow through the entire network.

\section{Experiments}

We compare both architectures across multiple scenarios: architectural components, performance on synthetic data, the Copiale manuscript, and sequence length variation.

\subsection{Architecture Comparison}

Both approaches employ CRNN-based visual encoding followed by attention-based decoding, but differ in key aspects. The transcription model uses a four-block CNN with 4-layer bidirectional LSTM, while Direct Image Decryption extends this with a deeper five-block CNN and 2-layer bidirectional LSTM. Both use 2-layer decoder LSTMs with 128-dimensional embeddings, though Direct Image Decryption employs 8-head multi-head attention compared to the simpler additive attention in the decryption stage.

The critical architectural distinction lies in the training paradigm. The two-stage pipeline trains two sequential models independently: transcription uses CTC loss to predict 132 cipher symbols, then decryption converts these discrete tokens to 62 plaintext characters using cross-entropy loss. This commits to discrete symbol decisions early, preventing gradient flow from decryption errors back to visual feature extraction. In contrast, Direct Image Decryption trains a single end-to-end model that directly predicts characters from images, maintaining continuous representations throughout and allowing visual features to adapt directly to decryption requirements through end-to-end gradient flow. Whether this end-to-end learning advantage outweighs architectural differences is the empirical question addressed in subsequent results. 
Comparison with the State-Of-The-Art could not be done, as currently baselines for a Direct Image Decryption approach do not exist. Work on this matter either delves into transcription or decryption separately; a conjoined baseline does not exist.

\subsection{Evaluation Metrics}
Models are evaluated using four case-insensitive metrics: \textbf{Token Accuracy} (percentage of exactly correct predictions, higher is better, [0.0--1.0]), \textbf{Normalized Edit Distance (NED)} (Levenshtein distance normalized by the length of the longer sequence, lower is better, [0.0--1.0]; equivalently, decryption success rate is reported as $1 - \text{NED}$), \textbf{Word Error Rate (WER)} (token-level error computed as $(S+D+I)/N$, lower is better, [0.0--1.0]), and \textbf{Character Error Rate (CER)} (character-level equivalent of WER, computed as $(S+D+I)/N$ where operations are counted over individual characters and $N$ is the number of characters in the reference; unlike NED, CER normalizes by reference length and penalizes insertions and deletions independently, lower is better, [0.0--1.0]).

\subsection{Results on Synthetic Data}

We evaluate both approaches in our generated synthetic data. All models were trained on the 115,000-image ``Faust'' dataset and tested on two datasets: the held-out ``Faust'' test set and the out-of-distribution Novalis dataset (1,300 images of Old German poetry).

\subsubsection{Transcription Performance}

\begin{figure}[ht]
\centering
\includegraphics[width=1\columnwidth]{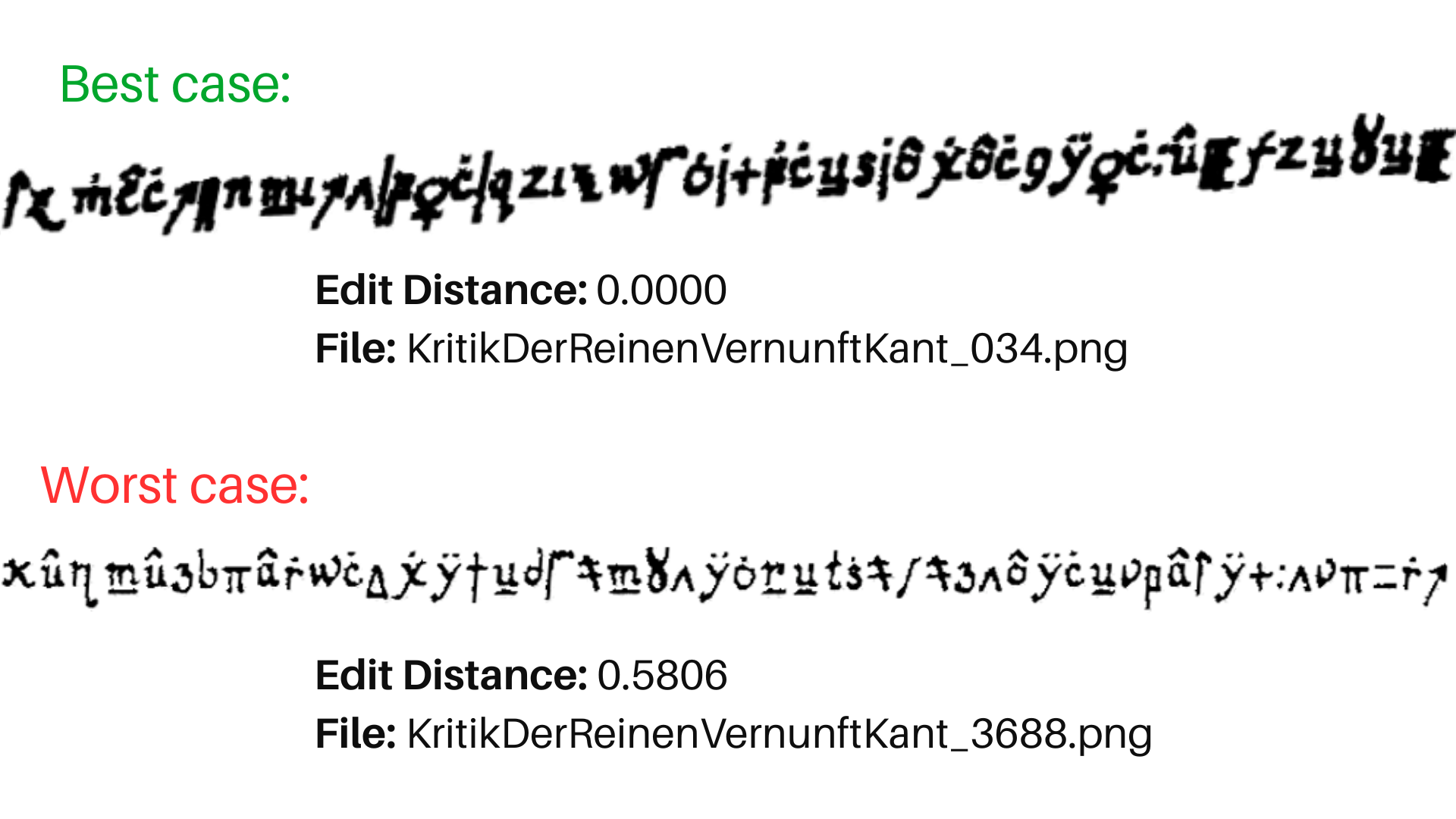}
\caption{Best and worst case examples of transcription on synthetic data.}
\label{fig:transcription-best-worst}
\end{figure}

In the traditional pipeline (namely 2-stage), the transcription model achieves strong performance on the ``Faust'' test set (Table~\ref{tab:transcription-synthetic}), with 91.5\% token accuracy and 4.3\% normalized edit distance. Figure~\ref{fig:transcription-best-worst} illustrates some of the best and worst case examples based on Edit Distance.

\begin{table}[ht]
\centering
\caption{Transcription performance on ``Faust''.}
\label{tab:transcription-synthetic}
\begin{tabular}{lc}
\hline
\textbf{Metric} & \textbf{Score} \\
\hline
Token Accuracy $\uparrow$ & 0.915 \\
Edit Distance $\downarrow$ & 0.043 \\
WER $\downarrow$ & 0.075 \\
CER $\downarrow$ & 0.076 \\
\hline
\end{tabular}
\end{table}

\subsubsection{Comparison: Two-Stage vs. Direct Image Decryption}

Table~\ref{tab:synthetic-comparison} shows the performance comparison of both approaches. On in-distribution data (Faust), Direct Image Decryption outperforms the two-stage pipeline across most metrics. The 1.1\% improvement in token accuracy validates our hypothesis that eliminating the transcription bottleneck reduces error propagation. The 49\% reduction in WER (0.206 $\rightarrow$ 0.105) demonstrates superior ability to generate coherent character sequences.

The performance gap widens substantially on the out-of-distribution Novalis dataset, which contains different vocabulary, sentence structures, and poetic phrasing compared to the training data. Direct Image Decryption achieves 75.8\% token accuracy compared to 69.5\%—a 6.3\% absolute improvement, suggesting that end-to-end learning enables better generalization. The two-stage pipeline's performance degradation (91.3\% $\rightarrow$ 69.5\%) is more severe than Direct Image Decryption's (92.4\% $\rightarrow$ 75.8\%), indicating that error propagation from transcription compounds when encountering unfamiliar text patterns. The WER reduction (59.7\% $\rightarrow$ 31.6\%) demonstrates that Direct Image Decryption maintains better sequence-level coherence even when faced with novel vocabulary and grammatical structures. These results provide strong evidence that the end-to-end paradigm offers robustness advantages beyond simple accuracy improvements.

\begin{table}[ht]
\centering
\caption{End-to-end decryption on synthetic data.}
\label{tab:synthetic-comparison}
\begin{tabular}{lccc}
\hline
\textbf{Dataset} & \textbf{Metric} & \textbf{2-Stage} & \textbf{Direct} \\
\hline
\multirow{4}{*}{\textbf{Faust}} 
& Token Acc. $\uparrow$ & 0.913 & \textbf{0.924} \\
& Edit Dist. $\downarrow$ & 0.045 & \textbf{0.038} \\
& WER $\downarrow$ & 0.206 & \textbf{0.105} \\
& CER $\downarrow$ & \textbf{0.056} & 0.065 \\
\hline
\multirow{4}{*}{\textbf{Novalis}} 
& Token Acc. $\uparrow$ & 0.695 & \textbf{0.758} \\
& Edit Dist. $\downarrow$ & 0.190 & \textbf{0.162} \\
& WER $\downarrow$ & 0.597 & \textbf{0.316} \\
& CER $\downarrow$ & 0.204 & \textbf{0.183} \\
\hline
\end{tabular}
\end{table}

\subsection{Results on the Copiale Manuscript}

Next, we evaluate both approaches on approximately 2,000 line images from the real 18th-century Copiale cipher \cite{Fornes:24}, probing model behavior under real conditions.

\subsubsection{Transcription Performance}

In the two-stage pipeline, the transcription component generalizes well to the original manuscript (Table~\ref{tab:transcription-copiale}), achieving 91.1\% token accuracy—only 0.4 percentage points lower than on synthetic data, confirming that visual recognition of Copiale glyphs transfers effectively.

\begin{table}[ht]
\centering
\caption{Transcription performance on Copiale.}
\label{tab:transcription-copiale}
\begin{tabular}{lc}
\hline
\textbf{Metric} & \textbf{Score} \\
\hline
Token Accuracy $\uparrow$ & 0.911 \\
Edit Distance $\downarrow$ & 0.023 \\
WER $\downarrow$ & 0.017 \\
CER $\downarrow$ & 0.014 \\
\hline
\end{tabular}
\end{table}

\subsubsection{End-to-End Decryption Performance}

In contrast to transcription, joint end-to-end decryption performance degrades substantially on the authentic manuscript (Table~\ref{tab:copiale-comparison}). As expected, both approaches perform significantly worse than on synthetic data, with absolute accuracies well below practical usability.

The two-stage pipeline achieves 39.6\% token accuracy, while Direct Image Decryption reaches 51.4\%—an absolute improvement of 11.8 percentage points (30\% relative). Direct Image Decryption also reduces WER from 89.0\% to 76.0\% and CER from 43.0\% to 39.3\%, indicating improved sequence-level coherence and fewer catastrophic decoding failures. Despite this improvement, both models fail quite often on the Copiale manuscript, demonstrating that historical cipher decipherment remains a challenge under current data constraints. The fact that transcription succeeds (91.1\%) while decryption fails (39.6-51.4\%) could suggest that the bottleneck lies in linguistic modeling rather than in visual recognition.

\begin{table}[ht]
\centering
\caption{End-to-end decryption on Copiale.}
\label{tab:copiale-comparison}
\begin{tabular}{lccc}
\hline
\textbf{Metric} & \textbf{2-Stage} & \textbf{Direct} & \textbf{$\Delta$} \\
\hline
Token Acc. $\uparrow$ & 0.396 & \textbf{0.514} & +11.8\% \\
Edit Dist. $\downarrow$ & 0.428 & \textbf{0.303} & -12.5\% \\
WER $\downarrow$ & 0.890 & \textbf{0.760} & -13.0\% \\
CER $\downarrow$ & 0.430 & \textbf{0.393} & -3.7\% \\
\hline
\end{tabular}
\end{table}

\subsubsection{Analysis: The Data Scarcity Problem}

The performance collapse from 91-92\% (synthetic) to 40-51\% (real) reveals a fundamental challenge: insufficient training data on real manuscripts. Critically, when we train models on reduced synthetic data—20,000 images yield ~53\% accuracy, 8,000 images yield ~31\% (Table~\ref{tab:linguistic-impact})—demonstrating that models require large-scale data to learn robust decipherment, regardless of whether data is synthetic or real.

This reframes our understanding of the synthetic-to-real gap. The problem is not primarily that synthetic data is qualitatively inadequate, but rather that we have \textbf{57 times less real data} (2,000 images) than synthetic training data (115,000 images). Models achieve 91-92\% accuracy on synthetic data because they have sufficient examples to learn robust patterns. When applied to real manuscripts, they fail because there was not enough real examples.
Performance degradation on limited data stems from:

\textbf{Insufficient statistical coverage}—with only 2,000 real manuscript images, models encounter symbol combinations and degradation patterns during testing never seen during training. Deep learning models require tens of thousands of examples to generalize robustly. Our ablation study presented in Table 5 confirms this: reducing synthetic training data by 82\% (115k $\rightarrow$ 20k) causes a 38+ percentage point accuracy drop, comparable to synthetic-to-real degradation. 

\textbf{Compounding error propagation}—when trained on limited data, both visual feature extraction and language modeling components underfit, leading to catastrophic failure cascades. 

\textbf{Linguistic domain specificity}—while the Copiale manuscript's esoteric content differs from our Faust/Kant/Bible/Nachsommer corpus, this vocabulary mismatch would be learnable given sufficient real manuscript data.

\begin{table}[ht] 
\centering 
\caption{Impact of training corpus and scale.} 
\label{tab:linguistic-impact} 
\begin{tabular}{lcc} 
\hline
\textbf{Corpus} & \textbf{Images} & \textbf{Accuracy $\uparrow$} \\ 
\hline
Faust & 115,000 & 92.4\% \\ 
American Psycho & 20,000 & 53.7\% \\ 
East of Eden & 8,000 & 31.7\% \\ 
Copiale Real & 2,000 & 51.4\% \\ 
\hline
\end{tabular}
\end{table}
Despite severe data scarcity, Direct Image Decryption's 11.8\% improvement over the two-stage baseline remains meaningful, validating that end-to-end learning reduces error propagation regardless of training set size. Critically, the advantage is most pronounced on challenging real-world data: 1.1\% improvement on abundant synthetic data versus 11.8\% improvement on scarce real data, suggesting that end-to-end gradient flow provides robustness benefits that become more valuable precisely when training data is limited.

The core challenge is clear: we need more real manuscript data. The 51\% token accuracy represents the performance ceiling achievable with 2,000 training examples. Scaling to 10,000-50,000 real manuscript images would likely yield higher accuracy needed for practical deployment. Until such data becomes available, automatic decipherment systems serve as tools for augmenting and aiding human expert analysis.

\begin{figure*}[ht] 
\centering 
\includegraphics[width=0.8\linewidth]{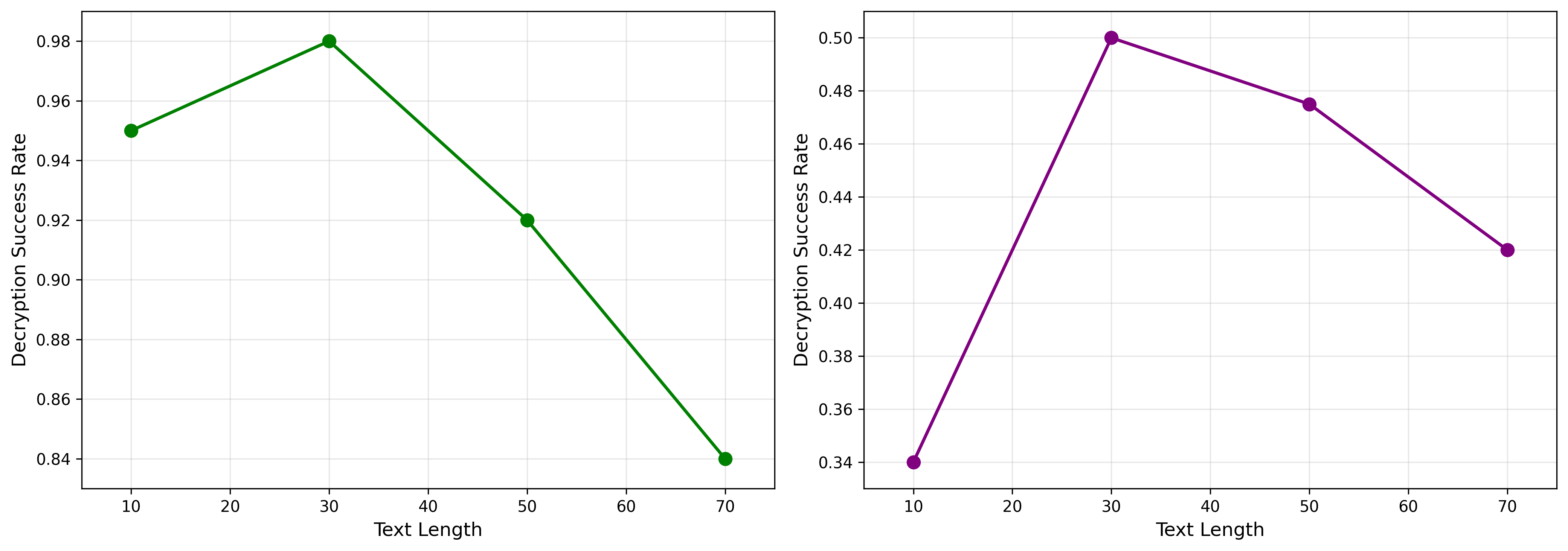} 
\caption{Decryption success rate relative to sequence length for synthetic ``Faust'' (left) and the original Copiale manuscript (right).} 
\label{fig:sequence-length-results} 
\end{figure*}

\subsection{Sequence Length Experiment}

How does sequence length affect decryption success? To answer this, we evaluated models on dataset variations categorized by length. For the ``Faust'' dataset: very short (3--12 characters), normal (12--40 characters), and long sequences (40--70 characters). As shown in Figure~\ref{fig:sequence-length-results}, decryption success rate (1-NED) remains high for shorter sequences but declines as text length increases, dropping from nearly 1.00 to approximately 0.84 for the 70-character range.

For the original Copiale manuscript, the model struggles significantly with extremely short inputs (below 10 tokens), where success rates hover around 0.35. However, performance peaks and stabilizes once the reference length exceeds 10 tokens, maintaining a success rate near 0.5 for standard manuscript line lengths.

The divergence suggests that while synthetic models may overfit to specific length patterns, real manuscript data requires a minimum threshold of linguistic context for reliable decryption. In synthetic data, increased complexity of longer strings drives errors up, whereas in Copiale, the primary hurdle is lack of sufficient information in very short snippets. This asymmetry reveals that the models learn different failure modes depending on their training distribution—synthetic models struggle with long-range dependencies, while real-data models need sufficient context to disambiguate the noisy visual features.

\section{Conclusions and Future Work}

This work has explored the transition from traditional multi-stage pipelines to end-to-end neural architectures for decipherment of historical encrypted manuscripts. By introducing Direct Image Decryption, we have demonstrated that mapping encrypted manuscript images directly to decrypted plaintext is both feasible and superior to the traditional transcription-decryption sequence.

For the evaluation, we have developed a comprehensive synthetic data generation pipeline producing over 115,000 realistic Copiale-like manuscript images from historical German texts. We implemented and evaluated a CRNN-based transcription model with CTC loss, achieving 91.5\% token accuracy on synthetic data and 91.1\% on the original Copiale manuscript. We then compared Direct Image Decryption against the traditional 2-step approach, demonstrating consistent improvements across all evaluated datasets.

Direct Image Decryption consistently outperformed the two-stage baseline by a mean of 6\% token accuracy (Faust, Novalis, Copiale), with advantages becoming more pronounced under challenging conditions—11.8\% improvement on the real Copiale versus 1.1\% on synthetic data. This validates our hypothesis that eliminating the intermediate transcription step significantly reduces error propagation, thus improving performance.

While transcription models generalize effectively to real manuscripts, both pipelines' decryption performance drops sharply, revealing a fundamental data scarcity challenge. Our analysis proves this is mainly quantitative rather than qualitative: models require many examples to learn robust linguistic patterns, yet we possess 57 times less real data (2,000 images) than synthetic data (115,000 images). The consistency of cipher glyphs allows visual recognition models trained on synthetic data to perform reliably on authentic 18th-century handwriting, suggesting the primary bottleneck lies in linguistic modeling rather than visual feature extraction.

These findings underscore that the path forward for automatic historical cipher decipherment lies not in algorithmic innovation alone, but in scaling real data. Until such data becomes available and the architecture is tested on other ciphers, systems like Direct Image Decryption serve as tools for augmenting human expert analysis, offering relative improvements that can meaningfully reduce the manual effort required for decipherment.

The main contribution of this work is to show that end-to-end image-to-plaintext modeling can help in the decipherment workflow, especially by reducing transcription-related error propagation and helping prioritize expert effort. However, our experiments remain restricted to a cipher with a known key, so the method should be understood not as a replacement for scholarly decipherment, but as a computational tool that may assist it.

Concerning future work, several research avenues remain to bridge the gap between experimental models and practical tools. First, one should explore generative or diffusion models for more realistic synthetic data generation, able to capture subtle characteristics like ink flow variations and aging patterns that current augmentation approximates but does not fully replicate. Second, our Direct Image Decryption architecture should be evaluated on other historical ciphers, such as the Borg or Ramanacoil manuscripts, to determine whether end-to-end mapping advantages generalize beyond other historical substitution-based ciphers. Finally, self-supervised learning could allow models to learn from untranscribed manuscripts, while AI-in-the-loop systems integrating expert feedback would refine linguistic priors and create high-quality training data through interactive correction, accelerating decipherment of unsolved historical manuscripts.

\section*{Acknowledgments}
This work has been partially supported by Riksbankens Jubileumsfond, grant M24-0028 (Echoes of History: Analysis and Decipherment of Historical Writings, DESCRYPT), the Spanish project PID2024-157778OB-I00 (SUKIDI) from the Ministerio de Ciencia e Innovación, the Departament de Cultura of the Generalitat de Catalunya, and the CERCA Program / Generalitat de Catalunya. Alicia Fornés acknowledges financial support for her general research activities from ICREA under the ICREA Academia (Departament de Recerca i Universitats de la Generalitat de Catalunya). Lei Kang acknowledges financial support from the Beatriu de Pinós del Departament de Recerca i Universitats de la Generalitat de Catalunya (2022 BP 00256).

\bibliographystyle{histocrypt}
\bibliography{histocryptbibliography}

\end{document}